\documentclass{esub2acm}
\begin{document}

\newcommand{\bequa}{\begin{equation}}
\newcommand{\eequa}{\end{equation}}
\newcommand{\noi}{\noindent}
\newcommand{\pdf}{\emph{pdf}}
\newcommand{\ip}{$\mathcal{IPERM}$}
\newcommand{\Remp}{$R_{emp}^{\mathbf{\Upsilon}_{N}}(h)$}
\newcommand{\hemp}{$h_{emp}^{\mathbf{\Upsilon}_{N}}$}
\newtheorem{defi}{Definition}
\newtheorem{theo}{Theorem}

\title{Modeling the Uncertainty in Complex\\ Engineering Systems}
\author{A. Aziz Guergachi\\
        Information Technology Group of ZENON Environmental, Inc.}

\begin{abstract}
Existing procedures for model validation have been deemed inadequate for many engineering systems. The reason of this inadequacy is due to the high degree of complexity of the physical mechanisms that govern these systems. It is proposed in this paper to shift the attention from modeling the engineering system itself to modeling the uncertainty that underlies its behavior. A mathematical framework for modeling the uncertainty in complex engineering systems is developed. This framework uses the results of computational learning theory. It is based on the premise that a system model is a learning machine.
\end{abstract}
\begin{bottomstuff}
\begin{authinfo}
This work was financially supported by CIDA and NSERC.
\end{authinfo}
\permission
\end{bottomstuff}
\markboth{A. Guergachi}
     {Modeling the Uncertainty in Complex Engineering Systems}
\maketitle
\section{Introduction}
Modeling of engineering systems such as wastewater treatment plants, groundwater contaminant transport, membrane fouling, sediment transport phenomena, $\cdots$ is traditionally carried out in three sequential steps:
\begin{itemize}
\item[i] \emph{model development}: the modeler collects the available knowledge about the studied system $S$ in the form of first principles, empirical laws and/or heuristic hypotheses. Based on this knowledge, the modeler develops a set of mathematical relationships (i.e., the system model $\mathcal{M}$) among the system state variables, which can generally be written in the form of a differential equation:
\begin{equation}
\dot{\mathbf{x}}= \mathbf{f}(t,\mathbf{x},\mathbf{p})  \label{equa2.1}
\end{equation}

where $t$ is the time, $\mathbf{x}$ is the system state vector, $\mathbf{p}$ is the model parameter vector and $\mathbf{f}$ is a mathematical function generally nonlinear.

\item[ii] \emph{model identification}: after the model is developed, the modeler uses a set $\Upsilon_N$ ($N \in \aleph^\circ$) of empirical data:
\begin{equation}
\Upsilon_N :\ \  \mathbf{x}^{data}(t_1), \mathbf{x}^{data}(t_2), \ldots, \mathbf{x}^{data}(t_N) \label{equa2.2}
\end{equation}
collected from the real operation of the system, to identify the model parameters. This step usually requires the minimization of an objective function $J(\mathbf{p})$ of the form:
\begin{equation}
J(\mathbf{p})=\sum_{k=1}^{N}\|\mathbf{x}(\mathbf{p},t_k)-\mathbf{x}^{data}(t_k)  \|^2 \label{equa2.3}
\end{equation}
where $\mathbf{x}(\mathbf{p},t)$ represents the solution to the model equation \ref{equa2.1}. In most cases, the data set $\Upsilon_N$ would actually be divided into two subsets $\Upsilon_{N_1}$ and $\Upsilon_{N_2}$ (\mbox{$N=N_1+N_2$}). The first subset (called identification sample) is used for the model parameter vector identification, and the second (called validation sample) for model validation (step below).
\item[iii] \emph{model validation}: in this step, the identified system model is tested on the validation subset $\Upsilon_{N_2}$ that it has never ``seen''. If the model performs well on this sample, then it is retained. Otherwise, the model structure is adjusted and the validation procedure repeated.
\\
\end{itemize}

The foregoing model validation procedure (called \emph{cross validation}) has been criticized in many areas of engineering. In wastewater engineering, for example, \citeN{jeppsson:wwt} pointed out that, \emph{``in strict sense, model validation is impossible''} with the existing validation techniques. Similarly, \citeN{zheng:contaminant} noted that, in groundwater engineering, \emph{``models, like any scientific hypothesis, cannot be validated in the absolute sense \ldots They can only be invalidated''}. \citeN{konikow:groundwater} suggested that terms like model verification and model validation convey a false sense of truth and accuracy and thus should be abandoned in favor of more realistic assessment descriptors such as history-matching and benchmarking.
\\

The engineering systems for which the cross validation procedure is deemed inadequate all share one same feature: the mechanisms that govern each one of them are so complex that no one model can be considered to describe these mechanisms in their entirety. The predictions of a model, no matter how sophisticated it is, are not guaranteed to match the reality. In this paper, it is proposed to shift the attention from modeling the system itself to modeling the uncertainty that underlies its behavior. The aim is to answer questions such as: what makes uncertainty high or low? How can it be controlled and to what extent can it be reduced?
\\

A mathematical framework for modeling the uncertainty in complex engineering systems is developed in this paper. This framework is based on the premise that a system model is \emph{learning machine}. The model identification procedure is viewed as a \emph{learning problem} or, equivalently, an \emph{information transfer} from a finite set of real data $\Upsilon_{N}$ into the system model.
\\

The framework of this paper is based on the extensive research work by \citeN{Vapnik:Estimation, Vapnik:nature,Vapnik:theory} and that of \citeN{VC:1,VC:2,VC:3} in the area of mathematical statistics and its applications to computational machine learning theory. The next section shows why and how a system model can be considered as learning machine. The remainder of the paper is devoted to the framework development.

\section{A System Model is a Learning Machine}
Assume that we are interested in the variations of one state variable $x_{i_0}$ of the system $S$ and consider the model differential equation that governs the dynamics of this variable:
\[
\dot{x}_{i_0} = f(t,\mathbf{x}, \mathbf{p})
\]
or
\bequa
\frac{dx_{i_0}}{dt} = f(t,\mathbf{x}, \mathbf{p})  \label{equa4.12}
\eequa
where $t$ is the time, $\mathbf{x}$ is the process state vector, $\mathbf{p}$ is the parameter vector and $f$ is a real-valued function. This equation represents one component of the vector differential equation:
\[\dot{\mathbf{x}} =  \mathbf{f}(t,\mathbf{x},\mathbf{p})  \]
of the system model $\mathcal{M}$. However, the vectors $\mathbf{x}$ and $\mathbf{p}$ in equation \ref{equa4.12} do not necessarily contain all of their components. Normally, they should be denoted as $\mathbf{x}_{x_{i_0}}$ and $\mathbf{p}_{x_{i_0}}$ and equation \ref{equa4.12} should become:
\bequa
\frac{dx_{i_0}}{dt} = f(t,\mathbf{x}_{x_{i_0}}, \mathbf{p}_{x_{i_0}})  \label{equa4.13}
\eequa
in order to highlight the fact that $\mathbf{x}$ and $\mathbf{p}$ contains only those state variables and parameters, respectively, that influence the dynamics of $x_{i_0}$.
\\

\noi This study will be limited to the  case of \emph{autonomous systems}, i.e., systems whose models do not depend explicitly on time. In other words, the general model equation that governs $x_{i_0}$ can be written as:
\bequa
\frac{dx_{i_0}}{dt} = f(\mathbf{x}_{x_{i_0}}, \mathbf{p}_{x_{i_0}})  \label{equa4.14}
\eequa

\noi In addition to $x_{i_0}$, all state variables, components of $\mathbf{x}_{x_{i_0}}$, are assumed to be directly and separately measurable.
\\

\noi Using the Euler method to numerically integrate equation \ref{equa4.14}, the time is discretized with a time step of $\Delta t$ and then $x_{i_0}$ is computed at times
\[ t_1=\Delta t\ , \ \ t_2=2 \, \Delta t\ ,\ \ \ldots\ , \ \ t_n=n \, \Delta t\ , \ \ \ldots\]
using the following equation:
\bequa
x_{i_0}(t_n) = x_{i_0}(t_{n-1}) + \Delta t \, f\left(\mathbf{x}_{x_{i_0}}(t_{n-1}\right), \mathbf{p}_{x_{i_0}}) \label{equa4.15}
\eequa
Define $w^{\mathcal{M}}$ as the value of $x_{i_0}$ to be predicted by the model $\mathcal{M}$, that is: \[w^{\mathcal{M}} = x_{i_0}(t_n)\]
\noi Similarly, define the vector $\mathbf{v}$ as:
\bequa
\mathbf{v} = [x_{i_0}(t_{n-1}), \mathbf{x}_{x_{i_0}}(t_{n-1})^T]^T \label{equa4.16}
\eequa
The superscript $\ ^T$ means transposed vector. The number $w^{\mathcal{M}}$ takes values from a sub-set $W$ of the real line $\Re$, and vector $\mathbf{v}$ from a multi-dimensional space $V$. 
\\

\noi Now introduce the real-valued function $H$ defined as:
\bequa
H(\mathbf{v},\mathbf{p}_{x_{i_0}})= x_{i_0}(t_{n-1}) + \Delta t \, f(\mathbf{x}_{x_{i_0}}(t_{n-1}), \mathbf{p}_{x_{i_0}}) \label{equa4.17}
\eequa
The expression of this function corresponds to that of the right-hand side of equation \ref{equa4.15}. The latter equation becomes then:
\bequa
w^{\mathcal{M}} = H(\mathbf{v},\mathbf{p}_{x_{i_0}}) \label{equa4.16}
\eequa
For a fixed parameter vector $\mathbf{p}_{x_{i_0}}$, $H(\ .\ , \mathbf{p}_{x_{i_0}})$ represents a mapping function from $V$ to $W$:
\bequa
\begin{array}{llll}  
H(\ .\ , \mathbf{p}_{x_{i_0}}): &  V   &  \rightarrow & W \\
                     & \mathbf{v}  & \mapsto  & w^{\mathcal{M}} = H(\mathbf{v},\mathbf{p}_{x_{i_0}})
\end{array}  
\eequa
The parameter vector $\mathbf{p}_{x_{i_0}}$ takes values from a multi-dimensional space denoted here as $\Gamma$. Define the functional set $\mathcal{H}_{\mathcal{M}}$ of all mappings $H(\ .\ , \mathbf{p}_{x_{i_0}})$ with $\mathbf{p}_{x_{i_0}} \in \Gamma$:
\bequa
\mathcal{H}_{\mathcal{M}} = \{H(\ .\ , \mathbf{p}_{x_{i_0}})\ |\ \  \mathbf{p}_{x_{i_0}} \in \Gamma \}
\eequa

\noi Now assume that a sequence of instances of the couple $(\mathbf{v},w)$:
\[ \Upsilon_N: \ \ (\mathbf{v}_1,w_1) , (\mathbf{v}_2,w_2) , \ldots , (\mathbf{v}_N,w_N)  \]
can be obtained from the real process operation, and consider an algorithm $\mathcal{A}$ that receives the sequence $\Upsilon_N$ as input and produces a parameter vector $(\mathbf{p}_{x_{i_0}})_{emp}$ corresponding to the function $H(\ .\ , (\mathbf{p}_{x_{i_0}})_{emp}) \in \mathcal{H}_{\mathcal{M}}$ that best approximates the real process response. In practice, this algorithm corresponds to the system model identification procedure which consists in minimizing an objective function of the form:
\bequa
J(\mathbf{p})= \sum _{k=1}^{N} |w_k - H( \mathbf{v}_k,\mathbf{p})|^2 \label{equa4.19}
\eequa
or, equivalently:
\bequa
R_{emp}(\mathbf{p})= \frac{1}{N}\sum _{k=1}^{N} |w_k - H( \mathbf{v}_k,\mathbf{p})|^2 \label{equa4.19}
\eequa
The subscript $\ _{emp}$ means ``empirical'' and the number $|w_k - H( \mathbf{v}_k,\mathbf{p})|^2$ represents a measure of the \emph{loss} between the desired response $w_k$ corresponding to the vector $\mathbf{v}_k$ and the model prediction represented by $H( \mathbf{v}_k,\mathbf{p})$.
\\

\noi A set of mapping functions equipped with an algorithm such as $\mathcal{A}$ is called a \emph{learning machine} in the area of artificial intelligence and computational learning theory. We have then shown above that the couple $\mathcal{LM}_{S}=(\mathcal{H}_{\mathcal{M}}, \mathcal{A})$, composed of a system model and an identification procedure, can be viewed as a learning machine. On the basis of this result, it is possible to develop a mathematical framework that will allow us to model the uncertainty that underlies the behavior of the engineering system $S$. The next sections of this paper are about the development of such framework.
\\

\noi \emph{Remark}: {\footnotesize Note that training of the machine \[\mathcal{LM}_{S}=(\mathcal{H}^{\mathcal{M}},\mathcal{A})\] associated with the system $S$ is carried out for a specific time $t_n$. This time is arbitrary, but fixed. The examples $(\mathbf{v}_{1},w_{1}),(\mathbf{v}_{2},w_{2}), \cdots, (\mathbf{v}_{N},w_{N})$ to be used for machine training should therefore correspond to a series of realizations of the system at time $t_n$. In practice, this is not possible, because the instance vector $\mathbf{v}$ and the outcome $w$ are measured only \textbf{once} at any time instant t. And what is obtained from these measurements is actually a \emph{time series}:
\[(\mathbf{v}_{t_1},w_{t_1}),(\mathbf{v}_{t_2},w_{t_2}), \cdots, (\mathbf{v}_{t_n},w_{t_n}),\cdots \]
whose terms represent the couples instance/outcome at successive time instants $t_1$, $t_2$, $\cdots$, $t_n$, $\cdots$. It corresponds to \emph{one realization} of the system $S$ in time. This realization would usually --- if not always --- be the only one that is available for investigating the system's behavior. The property that allows us to use the series $(\mathbf{v}_{t_i},w_{t_i})$ instead of $(\mathbf{v}_{i},w_{i})$ is called \emph{ergodicity}. This condition is quite weak and will be assumed to hold true for the studied system $S$. An extensive discussion of such condition can be found in \citeN{Guergachi:1}.} 

\section{General Description of the Framework}

In a certain environment $\mathcal{E}$, a situation $\mathbf{v}$ arises randomly and a transformer $\mathcal{T}$ acts and assigns to this situation $\mathbf{v}$ a number $w$ obtained as a result of the realization of a random trial. Formally, situation $\mathbf{v}$ represents a vector that takes values from an abstract space $V$ called \emph{instance space}. It is generated according to a \emph{fixed} but \emph{unknown} probability density function (\emph{pdf}) $P_{\mathbf{v}}$ defined on $V$. The number $w$, which is dependent on $\mathbf{v}$, takes values from another space $W\subseteq\Re$ called \emph{outcome space}. It is generated according to a conditional \emph{pdf} $P_{w|\mathbf{v}}$ defined on $W$, also \emph{fixed} but \emph{unknown}. The mathematical object $(\mathbf{v},w)$ arises then in the product space $Z=V \times W$ (called \emph{sample space}) according to the joint \emph{pdf} \mbox{$P_{(\mathbf{v},w)}=P_{\mathbf{v}}P_{w|\mathbf{v}}$}, which characterizes the probabilistic environment $\mathcal{E}$. In what follows, the couple $(\mathbf{v},w)$ is denoted as $z$ (to mean that it takes values from the sample space $Z$). Using this notation, the joint \pdf\  $P_{(\mathbf{v},w)}$ is then denoted as $P_{z}$. The vector $\mathbf{v}$ will be indifferently called ``situation'' or ``instance'' and the number $w$ ``outcome'' or ``transformer's response''.
\\

\noi If the behavior of transformer $\mathcal{T}$ is governed by a process which is a dynamic one, this transformer would usually possess several different \emph{operating modes}. To each mode would correspond a different \pdf\ $P_z$ and a different range of variation of $\mathbf{v}$ and $w$. To illustrate what is meant by ``operating mode'' here, consider for instance the behavior of an automotive engine: the operating conditions of such engine are not the same when the car is climbing a hill and when it is taking a highway. In the first case, the engine develop a very high torque and the speed is low, while in the second case, the same engine operates under opposite conditions: the speed is high but the torque is low. Another example that illustrates this concept of ``operating mode'' is a wastewater treatment plant using the activated sludge process: the operation of this plant can use little return of sludge and low solids in the aeration tank in order to achieve the objective of removing soluble substrate with relatively low oxygen supply. But this plant could also be operated with the purpose of aerobically destroying all of the organic solids in the waste, which can be done by returning all the sludge to the aeration tank. Thus, the same plant could operate under different operating conditions. In what follows, the operating mode of the transformer $\mathcal{T}$ will be denoted by $\mathcal{OM}$.
\\

\noi Associated with the environment $\mathcal{E}=(\mathcal{T},\mathcal{OM},z,P_z)$ is a \emph{learning machine} $\mathcal{LM}$ whose objective is to understand the behavior of the transformer $\mathcal{T}$. It receives a finite sequence $\mathbf{\Upsilon}_N$ of $N$ training examples: 
\[\mathbf{\Upsilon}_{N}:(\mathbf{v}_{1},w_{1}),(\mathbf{v}_{2},w_{2}), \ldots, (\mathbf{v}_{N},w_{N})\]
or, using the z-notation:
\[\mathbf{\Upsilon}_{N}:z_1, z_2, \ldots, z_N\]
generated and measured in the probabilistic environment $\mathcal{E}$ as a result of one \emph{realization} of this same environment. Based on these training examples, the learning machine $\mathcal{LM}$ selects a strategy that specifies the best approximation $w^{\mathcal{LM}}$ of the transformer's response for each instance $\mathbf{v}$. Once this strategy is selected, it will be used on \emph{all} future situations $\mathbf{v}$ arising in the environment $\mathcal{E}$, in order to predict the transformer's responses. This strategy, which is mathematically a mapping function from $V$ into $W$, is called a \emph{decision rule} and is chosen from a fixed functional space $\mathcal{H}$ called \emph{decision rule space}. 
\\

\noi The goal of $\mathcal{LM}$ is then to select, from the space $\mathcal{H}$, that particular decision rule which best approximates the transformer's response. The expression ``\emph{best approximation of the transformer's response}'' means ``\emph{closeness to the transformer's `general tendency'} $g^\mathcal{T}$''. The latter function is defined as follows:
\bequa
g^\mathcal{T}(\mathbf{v}) = \mathbf{E}(w \ |\ \mathbf{v}) = \int_W w\, P_{w|\mathbf{v}}(w\ |\ \mathbf{v})\, dw  \label{equa5.2}
\eequa
This function will be indifferently called `general tendency' or `response function'. Closeness is understood in the sense of the metric $\mathcal{D}$ defined in the following way:
\bequa
\forall h \in \mathcal{H},\ \  \mathcal{D}(h,g^\mathcal{T}) = \sqrt{\mathbf{E}\textbf{(}\ l(h(\mathbf{v}),g^\mathcal{T}(\mathbf{v}))\ \textbf{)}} = \sqrt{\int_V l(h(\mathbf{v}),g^\mathcal{T}(\mathbf{v}))\, P_{\mathbf{v}}(\mathbf{v})\, d\mathbf{v}} \label{equa5.1}
\eequa
where $l$ is defined throughout this paper as the quadratic loss:
\[\forall(a,b) \in \Re^2, \ \ \ l(a,b) = |a-b|^2 \]
\\

\noi After receiving the sequence $\Upsilon_N$ of training examples, the learning machine $\mathcal{LM}$ selects that particular decision rule $h_0$ that minimizes $\mathcal{D}(h,g^\mathcal{T})$ on the space $\mathcal{H}$ ($h$ designates an element of $\mathcal{H}$ and $g^\mathcal{T}$ the transformer's ``general tendency''). Formally, this means finding the minimum of the function:
\[\begin{array}{llll}
\mathcal{D}(\ .\ ,g^\mathcal{T}): & \mathcal{H} & \rightarrow & \Re+ \\
   & h & \mapsto & \mathcal{D}(h,g^\mathcal{T})
\end{array}
 \]
and the decision rule $h_0$ at which this minimum is attained. To do so, $\mathcal{LM}$ implements an algorithm $\mathcal{A}$ whose ultimate goal is to find $h_0$ on the basis of the finite sequence $\Upsilon_N$ of training examples.
\\

\noi Note that $w$ is related to $g^\mathcal{T}(\mathbf{v})$ through the following relationship:
\bequa
w = g^\mathcal{T}(\mathbf{v}) + \epsilon  \label{equa5.3}
\eequa
where $\epsilon$ is the noise associated with the probabilistic environment $\mathcal{E}$. By the properties of conditional expectation, it follows from \ref{equa5.3} that:
\bequa
\textbf{E}(\epsilon\ |\ \mathbf{v}) = 0  \label{equa5.4}
\eequa
\\

\noi {\footnotesize \emph{Remark:} The decision rule space $\mathcal{H}$ is considered to be indexed by a subset of $\Re^{n}$ for some $n \geq 1$, that is, there exist an integer $n \geq 1$ and a subset $T \subseteq \Re^{n}$, such that the space $\mathcal{H}$ can be expressed as follows: $\mathcal{H}=\{h_{\mathbf{p}}|\ {\mathbf{p}}\in T\}$. This is the case for most engineering systems.}
 
\section{Overcoming the First Obstacle in Minimizing the value of $\mathcal{D}$ over the space $\mathcal{H}$}

The objective of the learning machine $\mathcal{LM} = (\mathcal{H},\mathcal{A})$ is to minimize the distance $\mathcal{D}(h,g^\mathcal{T})$ over all the decision rule space $\mathcal{H}$. This distance involves two functions: $h$ and $g^\mathcal{T}$. The function $h$ is an element of the space $\mathcal{H}$ and, as such, it is well known to $\mathcal{LM}$: once the components of $\mathbf{v}$ are measured, the value of $h(\mathbf{v})$ is readily computable. The problem however is $g^\mathcal{T}$. Not only it is an unknown function and impossible to derive from first principles (recall that the systems we are dealing with are complex ones), but there is no operational way of getting even sample measurements or any empirical information about it. $g^\mathcal{T}$ is indeed buried in noise. What we can measure, with respect to the transformer's response, is the outcome $w$, and $w$ contains in it both the value of $g^\mathcal{T}$ and noise, all mixed up.
\\

\noi So how should $\mathcal{LM}$ proceed to minimize $\mathcal{D}(h,g^\mathcal{T})$, when the only information it can get is in the form of noise-corrupted measurements of the outcome $w$ and, of course, the instance $\mathbf{v}$? Theorem \ref{theo5.1} will be of great help. Before stating it, we need the following definition:
\\

\begin{defi}[Expected Risk]
\label{def5.1} Let $\mathcal{E}=(\mathcal{T},\mathcal{OM},z,P_{z})$ be a probabilistic environment and, associated with it, a learning machine $\mathcal{LM}=(\mathcal{H},\mathcal{A})$. Let $h\in \mathcal{H}$ be a decision rule. The expected risk $R(h)$ of $h$ is defined as the expected value of the random variable:
\[l(h(\mathbf{v}),w) = |h(\mathbf{v}) - w|^2\]
when the vector $z=(\mathbf{v},w)$ is drawn at random in the sample space $Z=V \times W$ according to the pdf $P_{z}=P_{(\mathbf{v},w)}$ corresponding to environment $\mathcal{E}$. Formally, it is:
\bequa
R(h)=\mathbf{E}\textbf{\emph{(}}\ l(h(\mathbf{v}),w)\  \textbf{\emph{)}}=\int_{V \times W}l(h(\mathbf{v}),w)\,P_{(\mathbf{v},w)}(\mathbf{v},w)\,d\mathbf{v}\, dw \label{equa5.5}
\eequa \\
\end{defi}
\noi Also, to simplify the notations, we need the following definition:
\\

\begin{defi}[Simplifying Notations]
\label{def5.1} Let $\mathcal{E}=(\mathcal{T},\mathcal{OM},z,P_{z})$ be a probabilistic environment and, associated with it, a learning machine $\mathcal{LM}=(\mathcal{H},\mathcal{A})$. For every decision rule $h \in \mathcal{H}$, we define the real-valued function $l_h$ on the sample space $Z=V\times W$ as follows:
\bequa
\forall (\mathbf{v},w) \in V\times W,\ \ l_{h}(\mathbf{v},w)=l(h(\mathbf{v}),w) \label{equa5.6}
\eequa \\
\end{defi}

\noi Hence, using the z-notation, equations \ref{equa5.6} and \ref{equa5.5} become:
\bequa
\forall z=(\mathbf{v},w) \in Z,\ \ \ l_{h}(z)=l(h(\mathbf{v}),w)
\eequa

\bequa
\forall h \in \mathcal{H},\ \ \ R(h)=\mathbf{E}(l_{h}(z))=\int_{Z}l_{h}(z)\,P_{z}(z)\,dz \label{equa5.80}
\eequa
\\

\begin{theo}[Transition $\mathcal{D}(h,g^\mathcal{T}) \longrightarrow R(h)$ ]
\label{theo5.1} Let $\mathcal{E}=(\mathcal{T},\mathcal{OM},z,P_{z})$ be a probabilistic environment and, associated with it, a learning machine $\mathcal{LM}=(\mathcal{H},\mathcal{A})$. Let $h_0 \in \mathcal{H}$ be a fixed decision rule. Then the function:
\[ h \mapsto \mathcal{D}(h,g^\mathcal{T})\]
is minimal at $h_0$ if and only if the function:
\[ h \mapsto R(h) \]
is minimal at $h_0$.
\end{theo}

\noi \textbf{Proof.} Using equation \ref{equa5.2}, it can be shown that the equality:
\bequa
R(h)=\int_{V\times W}[w-g^{\mathcal{T}}(\mathbf{v})]^2\,P_{(\mathbf{v},w)}(\mathbf{v},w)\,d\mathbf{v}\, dw + [\mathcal{D}(h,g^{\mathcal{T}})]^2  \label{equa5.81}
\eequa
holds true for all $h\in \mathcal{H}$. Since the integral $\int_{V\times W}[w-g^{\mathcal{T}}(\mathbf{v})]^2\,P_{(\mathbf{v},w)}(\mathbf{v},w)\,d\mathbf{v}\, dw$ is independent of $h$, it follows that $\mathcal{D}(h,g^{\mathcal{T}})$ is minimal if and only if $R(h)$ is minimal, and that both functions attain their minimum at the same function $h_0$.$\Box$
\\

\noi Theorem \ref{theo5.1} is very important in simplifying the learning problem $\mathcal{LM}$ is faced with. What it means is that minimizing $\mathcal{D}(h,g^\mathcal{T})$ or, equivalently, the square of it $[\mathcal{D}(h,g^\mathcal{T})]^2$ over $\mathcal{H}$ amounts to minimizing $R(h)$ over the decision rule space. Look at the expressions of these two functions $[\mathcal{D}(h,g^\mathcal{T})]^2$ and $R(h)$:
\bequa
[\mathcal{D}(h,g^\mathcal{T})]^2 = \mathbf{E}\textbf{(}\ l(h(\mathbf{v}),g^\mathcal{T}(\mathbf{v}))\ \textbf{)} \label{equa5.9}
\eequa
and
\bequa
R(h)=\mathbf{E}\textbf{(}\ l(h(\mathbf{v}),w)\  \textbf{)} \label{equa5.10}
\eequa
From these expressions, it can be seen that, in the course of minimizing $\mathcal{D}(h,g^\mathcal{T})$, theorem \ref{theo5.1} allows us to replace the \emph{unknown} and \emph{non-measurable} noise-free value $g^\mathcal{T}(\mathbf{v})$ by the \emph{measurable} noise-corrupted value $w$, without loosing information on that decision rule $h_0$ at which the minimum of $\mathcal{D}(h,g^\mathcal{T})$ is attained.
\\

\noi The following theorem will be helpful for system uncertainty model development:
\\

\begin{theo}[First Inequality]
\label{theo5.2} Let $\mathcal{E}=(\mathcal{T},\mathcal{OM},z,P_{z})$ be a probabilistic environment and, associated with it, a learning machine $\mathcal{LM}=(\mathcal{H},\mathcal{A})$. Then the inequality:
\bequa
[\mathcal{D}(h,g^\mathcal{T})]^2 \leq R(h) \label{equa5.10'}
\eequa
holds true for any rule $h \in \mathcal{H}$.
\end{theo}

\noi \textbf{Proof.} This inequality is a direct consequence of equality \ref{equa5.81}.$\Box$
\\

\section{Second Obstacle: $P_z$ is not known to $\mathcal{LM}$}

Theorem \ref{theo5.1} is still not enough for $\mathcal{LM}$ to proceed to the determination of the rule $h_0$ that minimizes $\mathcal{D}(h,g^\mathcal{T})$. This is because $R(h)$ is function of the \pdf\ $P_z$: this \pdf\  embodies all sources of uncertainty in the environment $\mathcal{E}$ and, as such, it is not known. The objective --- and the power --- of the framework developed here consists in avoiding any strong \emph{a priori} assumption regarding the sources of uncertainty in $\mathcal{E}$. Consequently, in what follows, $P_z$ is considered fixed but unknown.
\\

\noi Now, having taken this stand on $P_z$, we have to find a way of minimizing $R(h)$ on the basis of only a \emph{finite} number $N$ of training examples $z_1$, $z_2$, \ldots, $z_N$. How to do that? By introducing a principle called $\mathcal{I}$nductive $\mathcal{P}$rinciple of $\mathcal{E}$mpirical $\mathcal{R}$isk $\mathcal{M}$inimization ($\mathcal{IPERM}$). This principle has emerged in the mid-eighties as a result of an extensive research work by \citeN{Vapnik:Estimation, Vapnik:nature,Vapnik:theory} and that of \citeN{VC:1,VC:2,VC:3}.

\section{Inductive Principle of Empirical Risk Minimization}

Before we state the $\mathcal{IPERM}$, we need to define the meaning of \emph{empirical risk} of a decision rule:
\\

\begin{defi}[Empirical Risk]
\label{def5.3} Let $\mathcal{E}=(\mathcal{T},\mathcal{OM},z,P_{z})$ be a probabilistic environment and, associated with it, a learning machine $\mathcal{LM}=(\mathcal{H},\mathcal{A})$. Let $h\in \mathcal{H}$ be a decision rule and $\mathbf{\Upsilon}_{N}=(z_{1},z_{2}, \ldots, z_{N})$
a finite sequence of $N$ training examples generated and measured in the probabilistic environment $\mathcal{E}$ as a result of one realization of this same environment. The empirical risk $R_{emp}^{\mathbf{\Upsilon}_{N}}(h)$ of $h$ on the sequence $\mathbf{\Upsilon}_{N}$ is defined as the arithmetic mean of the sequence of numbers:
\[(l_{h}(z_{i}))_{i=1,2, \ldots, N}\]
that is: 
\bequa R_{emp}^{\mathbf{\Upsilon}_{N}}(h)=\frac{1}{N}\sum_{i=1}^{N}l_{h}(z_{i}) \label{equa5.11}
\eequa
\\
\end{defi}

\noi Having introduced the concept of empirical risk, we can now define what is meant by an uncertainty model:

\begin{defi}[Uncertainty Model]
\label{def5.1} Let $\mathcal{E}=(\mathcal{T},\mathcal{OM},z,P_{z})$ be a probabilistic environment and, associated with it, a learning machine $\mathcal{LM}=(\mathcal{H},\mathcal{A})$. Let $\mathbf{\Upsilon}_{N}$ be a finite sequence of $N$ training examples from the environment $\mathcal{E}$ and $\eta$ a fixed real number in the interval ]0,1[. Let \hemp\ be a decision rule at which the empirical risk \Remp\ reaches its minimum. An $\eta$-uncertainty model (or simply uncertainty model) of the transformer $\mathcal{T}$ is any inequality of the type:
\bequa
\mathcal{D}(h_{emp}^{\mathbf{\Upsilon}_{N}},g^{\mathcal{T}}) \leq \varphi(e_1, e_2, \ldots, e_l) \label{equaDefUncertaintyModel}
\eequa
that satisfy the following two conditions:
\begin{itemize}
\item \ inequality \ref{equaDefUncertaintyModel} holds true with a probability of at least $1-\eta$.
\item \ $e_1$, $e_2$, \ldots ,$e_l$ are a set of uncertainty control variables and $\varphi$ is a real-valued function of these variables that satisfy the following:
\end{itemize}
\bequa
\left\{
\begin{array}{c} 
\mbox{the variables}\ e_i\ \mbox{and the function}\ \varphi\  \\
\mbox{\mbox{are readily} determinable/computable}
\end{array}
\right.  \label{equa3.12}
\eequa
\\
\end{defi}

\noi Expected and empirical risks, $R(h)$ and $R_{emp}^{\mathbf{\Upsilon}_{N}}(h)$, may seem to introduce new concepts in this framework, but they are not if we go back to the concepts of probability theory. To see that, \underline{fix} a decision rule $h$ in the space $\mathcal{H}$. Since $z$ is a random variable, the number $l_{h}(z)$ is then also a random variable. Denote it as $\xi$, that is:
\[ \xi = l_{h}(z) \]
(Recall that $h$ is fixed) From probability theory, we know that there are two measures of the \emph{central tendency} of a random variable such as $\xi$:
\begin{itemize}
\item an \emph{empirical measure}: given a series of realizations $\xi_1$, $\xi_2$, \ldots, $\xi_N$ of the variable $\xi$, this measure is constructed by computing the arithmetic average $(\sum_i \xi_i)/N$ of this series.
\item a \emph{mathematical measure}: this measure is expressed in terms of the \pdf\ $P_{\xi}$ of $\xi$, that is: $\int \xi P_{\xi}(\xi)\,d\xi$. It is called expected value.
\end{itemize}
In this framework, $R_{emp}^{\mathbf{\Upsilon}_{N}}(h)$ represents the empirical measure of the central tendency of $\xi = l_{h}(z)$ and $R(h)$ represents the mathematical one. The former measure is \emph{approximate} but \emph{computable}, the latter is \emph{exact} but \emph{unknown}. Also, note that, under some conditions with respect to the dependency and heterogeneity of the realizations $\xi_i$, the empirical measure converges to the mathematical one when $N$ is made infinitely large \cite{White:1}. This is known as the \emph{Law of Large Numbers} in probability theory. Applying this law to the case of the expected and empirical risks, we get that $R_{emp}^{\mathbf{\Upsilon}_{N}}(h)$ converges (in probability) to $R(h)$ as $N$ is made infinitely large. That is:
\bequa
R_{emp}^{\mathbf{\Upsilon}_{N}}(h) \rightarrow R(h)\ \ \ \ \ \ \ \  as \ \ N \rightarrow \infty  \label{equa5.12}
\eequa
The reader should note a very important fact here: \emph{the convergence \emph{\ref{equa5.12}} is valid for a \underline{fixed} decision rule $h$ in the space $\mathcal{H}$}. This is called pointwise convergence, as opposed to another type of convergence (called uniform convergence) that is discussed briefly in the next sections. The term ``pointwise'' refers to the fact that the convergence \ref{equa5.12} occurs only for fixed points of $\mathcal{H}$ and not for all points of this space simultaneously.
\\

\noi Now, let's state the $\mathcal{IPERM}$. This principle consists in implementing the following two actions:
\begin{itemize}
\item\ \textbf{action 1}: \emph{replace} the expected risk $R(h)$ by the empirical risk $R_{emp}^{\mathbf{\Upsilon}_{N}}(h)$ computed on the basis of one training sequence $\mathbf{\Upsilon}_{N}$;
\item\ \textbf{action 2}: \emph{take} the decision rule $h_{emp}^{\mathbf{\Upsilon}_{N}}$ at which $R_{emp}^{\mathbf{\Upsilon}_{N}}(h)$ reaches its minimum as a good representation of the \emph{best} rule $h_{0}$ that minimizes the expected risk $R(h)$.
\end{itemize}
Therefore, the implementation of the $\mathcal{IPERM}$ comes down to minimizing the empirical risk $R_{emp}^{\mathbf{\Upsilon}_{N}}(h)$, instead of the expected one $R(h)$, over the space $\mathcal{H}$ and then choosing that decision rule $h_{emp}^{\mathbf{\Upsilon}_{N}}$ at which the minimum of $R_{emp}^{\mathbf{\Upsilon}_{N}}(h)$ is reached to describe the transformer's behavior. Engineering systems modelers (in various areas of engineering such as chemical, civil or environmental) have been using this procedure for system model identification for years. The reader may then wonder why we are developing a new mathematical framework, if all what we are going to do is to turn back to the traditional model identification procedure? What is the point?
\\

\noi This framework is not about inventing new procedures, but rationalizing existing ones and modeling the uncertainty that is associated with them. Engineering systems modelers have been using the traditional identification procedure without being aware of the transitions:
\bequa
\mathcal{D}(h,g^\mathcal{T}) \longrightarrow R(h) \longrightarrow R_{emp}^{\mathbf{\Upsilon}_{N}}(h) \label{equa5.13}
\eequa
Their decision to rely on empirical risk minimization may be explained by the fact that mechanistic models (mechanistic models as opposed to balck-box ones) are usually assumed to contain adequate \emph{a priori} information about the real system and, as a result, very little information would be lost in the transition:
\bequa
R(h) \longrightarrow R_{emp}^{\mathbf{\Upsilon}_{N}}(h)  \label{equa5.14}
\eequa
Now we know that this is not true for a complex system, since all existing models represent just a simplified picture of the real system behavior. If the sequence $\Upsilon_{N}$ is a finite one, then there is definitely a loss of information in the transition \ref{equa5.14}, that has always been ignored by engineering systems modelers. The aim of this framework is to rationalize and investigate the validity of this transition. First, we determine in what cases the replacement of $R(h)$ by $R_{emp}^{\mathbf{\Upsilon}_{N}}(h)$ can be legitimatized and, second, evaluate the loss of information that occurs in the course of this replacement. To do so, we need to examine the applicability of the $\mathcal{IPERM}$, for which Vapnik's results will be of great help.

\section{Applicability of the $\mathcal{IPERM}$}

In the transition:
\bequa
\mathcal{D}(h,g^\mathcal{T}) \longrightarrow R(h) \label{equa5.15}
\eequa
there is absolutely no information loss, in virtue of theorem \ref{theo5.1}. As a result, $R(h)$ can be considered as an exact measure of the performance of the decision rule $h$ when this rule is selected by $\mathcal{LM}$ as an approximation of $g^\mathcal{T}$. The transition that is problematic is the second one:
\[ R(h) \longrightarrow R_{emp}^{\mathbf{\Upsilon}_{N}}(h) \]
$R_{emp}^{\mathbf{\Upsilon}_{N}}(h)$ is indeed just an estimation of $R(h)$. Of course, one may argue that replacing $R(h)$ by $R_{emp}^{\mathbf{\Upsilon}_{N}}(h)$, as suggested in {action 1} of the $\mathcal{IPERM}$, can be legitimatized by the fact that, according to the Law of Large Numbers, $R_{emp}^{\mathbf{\Upsilon}_{N}}(h)$ becomes a perfect estimation of $R(h)$ when the size $N$ of the sequence $\mathbf{\Upsilon}_{N}$ is made infinitely large. But, this fact cannot be used to justify {action 2} of $\mathcal{IPERM}$. Here is indeed the problem:

\begin{quote}
As was done above, denote the decision rules that minimize $R(h)$ and \Remp\  as $h_0$ and $h_{emp}^{\mathbf{\Upsilon}_{N}}$, respectively. This is equivalent to write that:
\bequa
R_{emp}^{\mathbf{\Upsilon}_{N}}(h_{emp}^{\mathbf{\Upsilon}_{N}})=\inf_{h\in \mathcal{H}}R_{emp}^{\mathbf{\Upsilon}_{N}}(h)  \label{equa5.16}
\eequa
and
\bequa
R(h_{0})=\inf_{h\in \mathcal{H}}R(h)   \label{equa5.17}
\eequa
{Action 2} of the $\mathcal{IPERM}$ stipulates to take \hemp\ as a good representation of the best rule $h_0$. For this to be justified, we need to ensure that \hemp\ is very ``close'' to minimizing the expected risk $R(h)$ which is, as pointed out previously, {an exact measure of rule's performance} (meaning rule's closeness to $g^\mathcal{T}$ in the sense of $\mathcal{D}$). In more concrete terms, we need that the value $R(h_{emp}^{\mathbf{\Upsilon}_{N}})$ of the expected risk at $h_{emp}^{\mathbf{\Upsilon}_{N}}$ be close to the minimum one $R(h_{0})$, for $N$ sufficiently large. That is:
\bequa
R(h_{emp}^{\mathbf{\Upsilon}_{N}}) \rightarrow R(h_{0})\ \ \ \ \ \ \ \ as\ \ \  N \rightarrow \infty  \label{equa5.18}
\eequa
(convergence is understood in probability)
\end{quote}
It has been shown \cite{VC:3} that the pointwise convergence \ref{equa5.12} does not guarantee the one that is really required for the purpose of the $\mathcal{IPERM}$, i.e., convergence \ref{equa5.18}. In other words, it is possible that convergence \ref{equa5.12} be satisfied, but $R(h_{emp}^{\mathbf{\Upsilon}_{N}})$ remains always far from $R(h_0)$ --- even for large values of $N$ ---, meaning that $h_{emp}^{\mathbf{\Upsilon}_{N}}$ would never constitute a good approximation to the transformer's behavior. It is therefore important to verify whether the $\mathcal{IPERM}$ is applicable or not before using it in any learning problems.
\\

\noi Taking into consideration the foregoing comments, the following definition shall be adopted for the meaning of the applicability of the $\mathcal{IPERM}$:
\\

\begin{defi}[Applicability of the $\mathcal{IPERM}$]
\label{def5.4}
Let $\mathcal{E}=(\mathcal{T},\mathcal{OM},z,P_{z})$ be a probabilistic environment and, associated with it, a learning machine $\mathcal{LM}=(\mathcal{H},\mathcal{A})$. Let $\mathbf{\Upsilon}_{N}$ be a finite sequence of $N$ training examples from the environment $\mathcal{E}$ and let $h_{emp}^{\mathbf{\Upsilon}_{N}}$ and $h_{0}$ be two decision rules that minimize the risks \Remp\ and $R(h)$, respectively (refer to equations \ref{equa5.16} and \ref{equa5.17}). The $\mathcal{IPERM}$ is said to be applicable to $(\mathcal{E},\mathcal{LM})$ if, for any $\varepsilon >0$, the following equality holds true:
\bequa
\lim_{N \rightarrow \infty} \mathbf{Pr}\left(\sup_{h \in \mathcal{H}} \delta[R(h),R_{emp}^{\mathbf{\Upsilon}_{N}}(h)]> \varepsilon \right)=0 \label{equaUnif} 
\eequa
$\delta$ being a deviation measure defined on the real line.\\
\end{defi}

\noi Now that the applicability of \ip\ has been defined, we need to develop a simple method of verifying it. In the foregoing discussion, it has been pointed out that the pointwise convergence \ref{equa5.12} is not enough to guarantee the applicability of \ip. A more stringent condition regarding the empirical risk convergence needs to be imposed. \citeN{VC:3} have showed that, for \ip\ to be applicable, it is \emph{necessary} and \emph{sufficient} that the empirical risk \Remp\ converges \emph{uniformly} to the expected risk $R(h)$ over the whole space $\mathcal{H}$ (convergence is understood in probability). Mathematically, uniform convergence means that equation \ref{equaUnif} holds true. Intuitively, it means that, as $N$ is made infinitely large, the whole curve of \Remp\ converges to that of $R(h)$ over the space $\mathcal{H}$. In this presentation, the theoretical part of such questions will not be detailed. Instead, the reader is referred to Vapnik's book ``\emph{Statistical Learning Theory}'' [1998] for the details. In what follows, Vapnik's results are presented in a more practical fashion, allowing direct application to the cases under study in this paper (i.e., engineering systems). The mathematical rigor is, however, preserved throughout the whole presentation.
\\

\noi A criterion to verify the applicability of the \ip\ is not the only thing that is needed here. We also want to know how much information is lost when $R(h)$ is replaced by \Remp. Here again, to evaluate this information loss, we need to define a measure of the deviation between $R(h)$ and \Remp. For this purpose, two deviation relative measures are introduced:
\begin{itemize}
\item \emph{relative measure} $\delta_1$ defined by:
\bequa
\forall (a_{1},a_{2})\in \Re^{2},\ \ \  \delta_1[a_{1},a_{2}]=\frac{a_{1}-a_{2}}{\sqrt{a_{1}}}  \label{equa5.20}
\eequa
\item \emph{relative measure} $\delta_2$ defined by:
\bequa
\forall (a_{1},a_{2})\in \Re^{2},\ \ \  \delta_2[a_{1},a_{2}]=\frac{a_{1}-a_{2}}{a_{1}}  \label{equa5.21}
\eequa
\end{itemize}
Each one of these two measures will be associated with a different weak prior information about $(\mathcal{E},\mathcal{LM})$.
\\

\noi Using these measures, the following theorem \ref{theo5.2} defines sufficient conditions for the applicability of \ip\ and helps evaluate the loss of information that occurs when $R(h)$ is replaced by \Remp:
\\

\begin{theo}[Applicability of the $\mathcal{IPERM}$]
\label{theo5.3}
Let $\mathcal{E}=(\mathcal{T},\mathcal{OM},z,P_{z})$ be a probabilistic environment and, associated with it, a learning machine $\mathcal{LM}=(\mathcal{H},\mathcal{A})$. Let $\mathbf{\Upsilon}_{N}$ be a finite sequence of $N$ training examples from the environment $\mathcal{E}$ and $\eta$ a real number in the interval $]0,1[$. Let $\delta$ be one of the deviation measures $\delta_1$ or $\delta_2$. If it is possible to establish some $\mathcal{W}$eak $\mathcal{P}$rior $\mathcal{I}$nformation $\mathcal{WPI}$ about $(\mathcal{E},\mathcal{LM})$ and construct a function $\mathcal{C}$ dependent on $N$, the whole set $\mathcal{H}$, $\mathcal{WPI}$ and the number $\eta$ such that both statements 1 and 2 listed below hold true, then the $\mathcal{IPERM}$ is applicable to $(\mathcal{E},\mathcal{LM})$.
When such function:
\[\mathcal{C}=\mathcal{C}(N,\mathcal{H},\mathcal{WPI},\eta)\]
exists, the $\mathcal{IPERM}$ is said to be $\delta$-applicable to $(\mathcal{E},\mathcal{LM})$ with the bound \linebreak  $\mathcal{C}(N,\mathcal{H},\mathcal{WPI},\eta)$.

\begin{itemize}
\item \emph{\textbf{Statement 1}}: for any $\eta \in ]0,1[$, the inequality:
\[ \sup_{h \in \mathcal{H}}\delta[R(h),R_{emp}^{\mathbf{\Upsilon}_{N}}(h)] \leq \mathcal{C}(N,\mathcal{H},\mathcal{WPI},\eta)\] 
is satisfied with probability of at least $1-\eta$.

\item \emph{\textbf{Statement 2}}: when $\mathcal{H}$,$\eta$ and $\mathcal{WPI}$ are fixed, then:
\[\lim_{N \rightarrow \infty}\mathcal{C}(N,\mathcal{H},\mathcal{WPI},\eta) = 0\]
\end{itemize}
\end{theo}

\noi \textbf{Proof.} Let $\varepsilon>0$ and $\eta \in ]0,1[$ be two fixed numbers. From statement 2, we infer that:
\[ \exists N_0 \in \aleph,\  \forall N>N_0,\  \mathcal{C}(N,\mathcal{H},\mathcal{WPI},\eta)< \varepsilon \]
Then, from statement 1, we get that for $N>N_0$, the inequality:
\[\sup_{h \in \mathcal{H}}\delta[R(h),R_{emp}^{\mathbf{\Upsilon}_{N}}(h)]\leq \varepsilon \]
is satisfied with probability of at least $1-\eta$. That is:
\[ \mathbf{Pr}\left(\sup_{h \in \mathcal{H}} \delta[R(h),R_{emp}^{\mathbf{\Upsilon}_{N}}(h)]> \varepsilon \right) < \eta\]
Thus, we have shown that, for any $\varepsilon>0$:
\[\forall \eta \in ]0,1[,\ \exists N_0\in\aleph,\ \forall N>N_0,\ \mathbf{Pr}\left(\sup_{h \in \mathcal{H}} \delta[R(h),R_{emp}^{\mathbf{\Upsilon}_{N}}(h)]> \varepsilon \right) < \eta \]
which means, by definition, that:
\[
\lim_{N \rightarrow \infty} \mathbf{Pr}\left(\sup_{h \in \mathcal{H}} \delta[R(h),R_{emp}^{\mathbf{\Upsilon}_{N}}(h)]> \varepsilon \right)=0\ \ \ \ \ \ \ \ \ \ \ \hspace{1cm}\Box\]
\\

\noi Now recall that the objective of this study is to develop uncertainty models (see definition \ref{def5.1}) for complex engineering systems. The following theorem defines a way of developing such models:
\\

\begin{theo}[Uncertainty Model]
\label{theo5.4}
Let $\mathcal{E}=(\mathcal{T},\mathcal{OM},z,P_{z})$ be a probabilistic environment and, associated with it, a learning machine $\mathcal{LM}=(\mathcal{H},\mathcal{A})$. Let $\mathbf{\Upsilon}_{N}$ be a finite sequence of $N$ training examples from the environment $\mathcal{E}$ and $\eta$ a real number in the interval $]0,1[$. Let $\mathcal{WPI}$ be some weak prior information about $(\mathcal{E},\mathcal{LM})$ and \hemp\ a decision rule at which the empirical risk \Remp\ reaches its minimum.
\begin{itemize}
\item If the $\mathcal{IPERM}$ is $\delta_1$-applicable to $(\mathcal{E},\mathcal{LM})$ with the bound $\mathcal{C}(N,\mathcal{H}, \mathcal{WPI},\eta)$, then the inequality:
\begin{eqnarray}
\hspace{-2cm}[\mathcal{D}(h_{emp}^{\mathbf{\Upsilon}_{N}},g^{\mathcal{T}})]^2 & \leq & R_{emp}^{\mathbf{\Upsilon}_{N}}(h_{emp}^{\mathbf{\Upsilon}_{N}}) +  \nonumber \\
 & &  \frac{\mathcal{C}^2(N,\mathcal{H},\mathcal{WPI},\eta)}{2} \left(1 + \sqrt{1+\frac{4\,R_{emp}^{\mathbf{\Upsilon}_{N}}(h_{emp}^{\mathbf{\Upsilon}_{N}})}{\mathcal{C}^2(N,\mathcal{H},\mathcal{WPI},\eta)}} \right) \label{equa5.22}
\end{eqnarray}  
holds true with probability of at least $1-\eta$.
\item If the $\mathcal{IPERM}$ is $\delta_2$-applicable to $(\mathcal{E},\mathcal{LM})$ with the bound $\mathcal{C}(N,\mathcal{H},\eta, \mathcal{WPI})$, then the inequality:
\bequa
[\mathcal{D}(h_{emp}^{\mathbf{\Upsilon}_{N}},g^{\mathcal{T}})]^2 \leq \frac{R_{emp}^{\mathbf{\Upsilon}_{N}}(h_{emp}^{\mathbf{\Upsilon}_{N}})}{(1-\mathcal{C}(N,\mathcal{H},\mathcal{WPI},\eta))_+}   \label{equa5.23}
\eequa  
holds true with probability of at least $1-\eta$, where $(a)_+ = \sup(a,0)$.
\end{itemize}
\end{theo}

\noi \textbf{Proof.} If the $\mathcal{IPERM}$ is $\delta_1$-applicable to $(\mathcal{E},\mathcal{LM})$ with the bound $\mathcal{C}(N,\mathcal{H},\eta, \mathcal{WPI})$, then, from theorem \ref{theo5.3}, it follows that (all inequalities hold with probability of at least $1-\eta$):
\[\frac{R(h_{emp}^{\mathbf{\Upsilon}_{N}})-R_{emp}^{\mathbf{\Upsilon}_{N}}(h_{emp}^{\mathbf{\Upsilon}_{N}})}{\sqrt{R(h_{emp}^{\mathbf{\Upsilon}_{N}})}} \leq \mathcal{C}(N,\mathcal{H},\mathcal{WPI},\eta) \]
Hence:
\begin{eqnarray}
\hspace{-2cm}R(h_{emp}^{\mathbf{\Upsilon}_{N}}) & \leq & R_{emp}^{\mathbf{\Upsilon}_{N}}(h_{emp}^{\mathbf{\Upsilon}_{N}}) +  \nonumber \\
 & &  \frac{\mathcal{C}^2(N,\mathcal{H},\mathcal{WPI},\eta)}{2} \left(1 + \sqrt{1+\frac{4\,R_{emp}^{\mathbf{\Upsilon}_{N}}(h_{emp}^{\mathbf{\Upsilon}_{N}})}{\mathcal{C}^2(N,\mathcal{H},\mathcal{WPI},\eta)}} \right) \nonumber 
\end{eqnarray}
and then, from theorem \ref{theo5.2}, it follows that:  
\begin{eqnarray}
\hspace{-2cm}[\mathcal{D}(h_{emp}^{\mathbf{\Upsilon}_{N}},g^{\mathcal{T}})]^2 & \leq & R_{emp}^{\mathbf{\Upsilon}_{N}}(h_{emp}^{\mathbf{\Upsilon}_{N}}) +  \nonumber \\
 & &  \frac{\mathcal{C}^2(N,\mathcal{H},\mathcal{WPI},\eta)}{2} \left(1 + \sqrt{1+\frac{4\,R_{emp}^{\mathbf{\Upsilon}_{N}}(h_{emp}^{\mathbf{\Upsilon}_{N}})}{\mathcal{C}^2(N,\mathcal{H},\mathcal{WPI},\eta)}} \right) \nonumber 
\end{eqnarray}  

\noi Similarly, if the $\mathcal{IPERM}$ is $\delta_2$-applicable to $(\mathcal{E},\mathcal{LM})$ with the bound $\mathcal{C}(N,\mathcal{H},\eta, \mathcal{WPI})$, then, from theorem \ref{theo5.3}, it follows that:
\[\frac{R(h_{emp}^{\mathbf{\Upsilon}_{N}})-R_{emp}^{\mathbf{\Upsilon}_{N}}(h_{emp}^{\mathbf{\Upsilon}_{N}})}{R(h_{emp}^{\mathbf{\Upsilon}_{N}})} \leq \mathcal{C}(N,\mathcal{H},\mathcal{WPI},\eta) \]
and then:
\[
[\mathcal{D}(h_{emp}^{\mathbf{\Upsilon}_{N}},g^{\mathcal{T}})]^2 \leq R(h_{emp}^{\mathbf{\Upsilon}_{N}}) \leq \frac{R_{emp}^{\mathbf{\Upsilon}_{N}}(h_{emp}^{\mathbf{\Upsilon}_{N}})}{(1-\mathcal{C}(N,\mathcal{H},\mathcal{WPI},\eta))_+}\ \ \ \ \ \ \ \ \Box
\]
\\

\noi The bound on the squared distance $[\mathcal{D}(h_{emp}^{\mathbf{\Upsilon}_{N}},g^{\mathcal{T}})]^2$, when it exists, is called \emph{guaranteed deviation} between $h_{emp}^{\mathbf{\Upsilon}_{N}}$ and $g^{\mathcal{T}}$, and denoted as $\varphi$ or as \[\varphi(N,\mathcal{H},R_{emp}^{\mathbf{\Upsilon}_{N}}(h_{emp}^{\mathbf{\Upsilon}_{N}}),\mathcal{WPI},\eta)\]

\section{The Vapnik-Chervonenkis (VC) Dimension}

One of the objects which the guaranteed deviation $\varphi$ is dependent on is the whole set $\mathcal{H}$ of decision rules. Now we need to know exactly what characteristic of $\mathcal{H}$ affects $\varphi$ and the uncertainty models \ref{equa5.22} and \ref{equa5.23}. Intuitive analysis of uncertainty in engineering systems shows that this characteristic is the complexity of $\mathcal{H}$ \cite{Guergachi:1}. The objective of this section is to define a measure of this complexity. This measure is known as the \emph{Vapnik-Chervonenkis dimension}, or simply VC dimension, named in honor of its originators, \citeN{VC:1}. The definition of this dimension is quite difficult to assimilate from the first reading. Because of this, an intuitive interpretation of VC dimension will be first given and, at the end of this section, a series of illustrative examples will be presented.

\subsection{Intuitive Introduction}

Consider the following concrete example:
\begin{itemize}
\item $V_1 = \Re$ and $W_1=\Re$;
\item $\mathcal{H}=\mathcal{H}_{line}$ is the set of all functions $h$ from $V$ into $W$ such that:
\[ \forall x \in V, \ \ \ h(x) = p_1x+p_2 \]
with $\mathbf{p}=(p_1,p_2) \in \Re^2$ is the parameter vector. 
\end{itemize}
If we had to assign a number to the complexity of this set of functions, then intuitively the number two, corresponding to the number of parameters, would be the most suitable one. Consider now this second example:
\begin{itemize}
\item $V_2 = \Re$ and $W_2=\Re$;
\item $\mathcal{H}=\mathcal{H}_{sine}$ is the set of all functions $h$ from $V$ into $W$ such that:
\[ \forall x \in V, \ \ \ h(x) = p_1 \sin(p_2x) \]
with $\mathbf{p}=(p_1,p_2) \in \Re^2$ is the parameter vector. 
\end{itemize}
Since the number of parameters that define this set is also two, we may be tempted to again assign the number two to the complexity of this set. If we do so, it would mean that $\mathcal{H}_{line}$ and $\mathcal{H}_{sine}$ have the same degree of complexity, which is obviously not correct: the set $\mathcal{H}_{line}$ is a family of just straight lines, while $\mathcal{H}_{sine}$ is a complex family of curves that can take many different shapes. The ``expressive power'' of $\mathcal{H}_{sine}$ is indeed much higher than that of $\mathcal{H}_{line}$. As a result, it should be expected that the complexity of $\mathcal{H}_{sine}$ be much higher than that of $\mathcal{H}_{line}$, and that is what we get when we consider the VC dimension as a measure of the complexity of the decision rule space.
\\

\noi Intuitively, the VC dimension may be considered as equal to the maximum number of points that the curves representing the functions of the decision rule space can pass through simultaneously. Straight lines (functions defined by $h(x)= p_1x + p_2$, space $\mathcal{H}_{line}$) can pass through any 2 points, but not any 3 points. Parabolas (functions defined by $h(x)= p_1x^2 + p_2x + p_3$, space $\mathcal{H}_{parab}$) can pass through any 3 points, but not any 4 points. Sine functions ($h(x)=p_1\sin(p_2x)$, space $\mathcal{H}_{sine}$) can pass through any number of points. Hence, if the VC dimension of a space $\mathcal{H}$ is denoted as $q(\mathcal{H})$, then:
\[ q(\mathcal{H}_{line}) = 2 \]
\[ q(\mathcal{H}_{parab}) = 3 \]
\[ q(\mathcal{H}_{sine}) = \infty \]

\noi The foregoing intuitive interpretation of VC dimension is approximate. A more precise definition of it is given in the next section.

\subsection{Definitions}

\noindent For every set $I$, the notation $2^I$ will designate the set of all subsets of $I$.

\begin{defi}[VC Dimension of a Family of Sets]
\label{def-6}
Let $\textbf{G}$ be some space ($\Re^n$ with $n>0$ for example or any other space). Let $\mathcal{G}$ be a family of subsets of $\textbf{G}$ (examples of $\mathcal{G}$ in the case of $\textbf{G}=\Re^2$ are the family of all open (or closed) balls of $\Re^2$ or the family of all half planes of $\Re^2$) and $I$ a finite subset of $\textbf{G}$. Let $\Pi^{\mathcal{G}}(I)$ be the subset of $2^I$ defined as follows:
\[ \Pi^{\mathcal{G}}(I)=\{\Lambda \in 2^I |\exists F\in \mathcal{G}, \Lambda=F\cap I \} \]
The finite set I is said to be \textbf{\emph{shattered}} by the family of sets $\mathcal{G}$ if \underline{$\Pi^{\mathcal{G}}(I)=2^I$}. The largest integer $q$ such that some finite subset $I\subset \textbf{G}$ of \textbf{\emph{size}} $q$ is shattered by $\mathcal{G}$ is called the \emph{Vapnik-Chervonenkis dimension} (VC dimension) of the family $\mathcal{G}$. It is denoted by $q=q(\mathcal{G})$. If such integer $q$ does not exist, then the VC dimension of $\mathcal{G}$ is said to be infinite.
\end{defi}

\begin{defi}[VC Dimension of a Family of Functions]
\label{def-7}
Let $\mathcal{F}$ be a family of real-valued functions on some space $\textbf{G}$ and $I$ a finite subset of $\textbf{G}$. For every function $f\in \mathcal{F}$, define the subset pos(f) of the space $\textbf{G}$ as follows:
\[ pos(f)=\{a\in \textbf{G}|\ f(a)>0\}\]
Then define the family $pos(\mathcal{F})$ of subsets of $\textbf{G}$ as follows:
\[pos(\mathcal{F})= \{pos(f)|\ f\in \mathcal{F}\} \]
The finite set $I$ is said to be shattered by the family of real-valued functions $\mathcal{F}$, if it is shattered by the family of subsets $pos(\mathcal{F})$. The \emph{Vapnik-Chervonenkis dimension} (VC dimension) $q(\mathcal{F})$ of the family $\mathcal{F}$ of real-valued functions is, by definition, equal to the \emph{Vapnik-Chervonenkis dimension} of the family of subsets $pos(\mathcal{F})$:
\[q(\mathcal{F})=q(pos(\mathcal{F}))\]\\
\end{defi}

\noi The VC dimension is then a purely combinatorial concept that has, \emph{a priori}, no connection with the geometric notion of dimension. In most situations, it is difficult to evaluate the VC dimension by analytic means. Usually, all what it is possible is to determine a bound on the VC dimension, that is, establish an inequality of the form: $q(\mathcal{F}) \leq q_0$ ($q_0 \in \aleph$). Also in some cases the VC dimension is simply approximated by the free parameters of the family $\mathcal{F}$. The following theorem shows how to determine it in some particular cases. It also establishes a link with the geometric notion of dimension.
\\

\begin{theo}[VC Dimension and Vector Space]
\label{theo5.5}
Let $\mathcal{F}$ be a family of real-valued functions on some space $\textbf{G}$. Fix any function $f_0$ from $\textbf{G}$ into $\Re$ and let $\mathcal{F}_0$ be the new family of functions defined by $\mathcal{F}_0=f_0+\mathcal{F}=\{f_0+f|\ f\in \mathcal{F} \}$. If $\mathcal{F}$ is an m-dimensional real vector space, then the VC dimension  $q(\mathcal{F}_0)$ of $\mathcal{F}_0$ is equal to $m$:
\[q(\mathcal{F}_0)=m\]
\end{theo}

\noi \textbf{Proof.} Refer to \cite{wenocur:1} for the proof of this theorem $\Box$.
\\

\subsection{Examples}

\begin{itemize}
\item \textbf{Example 1:} Consider the family of functions $h_{\mathbf{p}}$ defined from the space $\mathbf{G} = \Re^n$ ($n \in \aleph^\circ$ ) into $\{0,1\}$ by:
\[\forall \textbf{x}=(x_1,x_2, \ldots, x_n)\in \Re^n,\ \ \ \ h_{\mathbf{p}}(\textbf{x}) = \psi(\sum_{i=1}^n p_i x_i)  \]
where $\mathbf{p}=(p_1, p_2, \ldots, p_n, \theta)\in \Re^{n+1}$ is the parameter vector and $\psi$ is defined by (real threshold $\theta$):
\[\psi(a) = \left\{ \begin{array}{ll}
1   & \mbox{if $a \geq \theta$} \\
0   & \mbox{if $a < \theta$}
\end{array}
\right. \]
This family of functions is known as the \emph{perceptron} and is used in pattern recognition. Its VC dimension is equal to $n+1$ \cite{anthony:1}.

\item \textbf{Example 2:} Consider the family of real-valued functions $h_{\mathbf{p}}$ defined on some space $\mathbf{G}$ by:
\[\forall \mathbf{x} \in \mathbf{G},\ \ \ \ h_{\mathbf{p}}(\mathbf{x}) = \sum_{i=1}^n p_i \psi_i(\mathbf{x})  \]
where $\mathbf{p}=(p_1, p_2, \ldots, p_n)\in \Re^{n}$ is the parameter vector and $\psi_1$, $\psi_2$, \ldots, $\psi_n$ is a sequence of $n$ linearly independent real-valued functions. The VC dimension of this family of functions is equal to $n$ \cite{Vapnik:Estimation}. Note that the determination of this VC dimension results directly from theorem \ref{theo5.5}.

\item \textbf{Example 3:} Consider the family of functions $h_{\mathbf{p}}$ defined on $\mathbf{G}=\Re^2$ by:
\[\forall (x,y) \in \Re^2,\ \ \ \  h_{\mathbf{p}}(x,y) = (y - poly_n(x,\mathbf{p}))^2 \]
where $\mathbf{p}=(p_0,p_1, p_2, \ldots, p_n)\in \Re^{n+1}$ is the parameter vector and $poly_m(x,\mathbf{p})$ is a polynomial function of degree $n$ defined by:
\[\forall x \in \Re,\ \ \ \ poly_n(x,\mathbf{p})= p_0 + p_1 x + p_2 x^2 + \ldots + p_n x^n \]
The VC dimension of this family of functions $h_{\mathbf{p}}$ is at most $2n + 2$ \cite{Vapnik:nature}.

\item \textbf{Example 4:} Consider the family of functions $h_{\mathbf{p}}$ defined on $\mathbf{G}=\Re$ by:
\[ \forall x \in \Re, \ \ \ \ h_{\mathbf{p}}(x) = p_1 \sin(p_2x) \]
where $\mathbf{p}=(p_1, p_2)\in \Re^{2}$ is the parameter vector. The VC dimension of this family of functions is infinite \cite{Vapnik:theory}.
\end{itemize}

\noi From these examples, it can be seen that, generally speaking, the VC dimension of a family of functions is not always related to the number of parameters. It can be larger (example 4), equal (examples 1 and 2) or smaller (see \cite{Vapnik:nature} where new types of learning machines were constructed) than the number of parameters.

\section{VC dimension and applicability of the \ip}

In section 7, the concept of applicability of \ip\ and that of guaranteed deviation between the decision rule $h_{emp}^{\mathbf{\Upsilon}_{N}}$ that minimizes the empirical risk and the transformer's response function $g^{\mathcal{T}}$ were introduced. However, no methodology has been developed to determine the expression of the function $\mathcal{C}=\mathcal{C}(N,\mathcal{H},\mathcal{WPI},\eta)$ (see theorems \ref{theo5.3} and \ref{theo5.4}), which is the key function in implementing those concepts. In this section, some fundamental results with respect to the determination of such function are presented. These results make use of the VC dimension concept defined in the previous section and they are due to \cite{Vapnik:theory}. Extensive discussion and application of these results to model identification and quality evaluation can be found in \citeN{Guergachi:1}.
\\

\noi Before stating these results, we need to define a new space $l_\mathcal{H}$ and five different conditions. 
\\

\begin{defi}[Space $l_\mathcal{H}$]
\label{def5.7}
Let $\mathcal{E}=(\mathcal{T},\mathcal{OM},z,P_{z})$ be a probabilistic environment and, associated with it, a learning machine $\mathcal{LM}=(\mathcal{H},\mathcal{A})$. For every decision rule $h \in \mathcal{H}$ and a real number $\beta \in \Re^+$, we define the real-valued functions $l_{h,\beta}$ on the sample space $Z=V\times W$ as follows:
\[\forall z \in Z,\ \ \ l_{h,\beta}(z)=l_{h}(z)-\beta \]
The functional space of all functions $l_{h,\beta}$ will be denoted by $l_{\mathcal{H}}$:
\[ l_{\mathcal{H}}=\{l_{h,\beta}|\ (h,\beta)\in \mathcal{H} \times \Re^+\} \]\\
\end{defi}
\noi Now let's define the following conditions \textbf{\emph{C.1}}, \textbf{\emph{C'.1}}, \textbf{\emph{C.2}}, \textbf{\emph{C.3}} and \textbf{\emph{C'.3}}:
\\

\noi \textbf{\emph{C.1}\ \ \ $\mathcal{W}$eak $\mathcal{P}$rior $\mathcal{I}$nformation (1):}\\ 
\emph{There exists a positive number $M \in\, ]0,+\infty[$ such that:}
\[\sup_{h \in \mathcal{H},z \in Z} l_h(z) = M \] 
\\
\noi \textbf{\emph{C'.1}\ \ \ $\mathcal{W}$eak $\mathcal{P}$rior $\mathcal{I}$nformation (2):}\\ 
\emph{There exist a pair $(s,\tau)\in \Re^2$ with $s > 2$ and $\tau < +\infty$ such that:}
\[\sup_{h\in \mathcal{H}}\frac{\mathbf{E}^{1/s}([l_{h}(z)]^{s})}{R(h)} <  \tau \]
\\
\noi \textbf{\emph{C.2}\ \ \ VC Dimension:}\\ 
\emph{ The VC dimension $q=q(l_{\mathcal{H}})$ of the functional space $l_{\mathcal{H}}$ is finite.}
\\

\noi \textbf{\emph{C.3}\ \ \ \emph{i.i.d.} condition:}\\
\emph{The training examples:}
\[z_{1},z_{2}, \ldots, z_{N}\]
\emph{of the sequence $\mathbf{\Upsilon}_{N}$ are independent and identically distributed (i.i.d.).}
\\

\noi \textbf{\emph{C'.3}\ \ \ Weaker \emph{i.i.d.} condition:}\\
\emph{The real-valued random variables:}
\[l_{h}(z_{1});\ l_{h}(z_{2});\  \ldots;\  l_{h}(z_{N})\]
\emph{obtained by computing the values of $l_{h}$ at each one of the training examples $z_{i}$ of the sequence $\mathbf{\Upsilon}_{N}$, are independent and identically distributed (i.i.d.) for any $h\in \mathcal{H}$.} 
\\

\begin{theo}[$\mathcal{IPERM}$ applicability and VC (1)]
\label{theo5.6}
Let $\mathcal{E}=(\mathcal{T},\mathcal{OM},z,P_{z})$ be a probabilistic environment and, associated with it, a learning machine $\mathcal{LM}=(\mathcal{H},\mathcal{A})$. Let $\mathbf{\Upsilon}_{N}$ be a finite sequence of $N$ training examples from the environment $\mathcal{E}$ and $\eta$ a real number in the interval $]0,1[$. If the conditions \textbf{C.1}, \textbf{C.2} and \textbf{C.3} are satisfied, then the $\mathcal{IPERM}$ is $\delta_1$-applicable to $(\mathcal{E},\mathcal{LM})$ with the bound:
\bequa
\mathcal{C}=\sqrt{M\, \zeta}  \label{equa5.24}
\eequa
where:
\begin{itemize}
\item The number $\zeta$ is:
\[\zeta=4 \frac{\left[ q \left( \ln \left( \frac{2N}{q}\right) +1\right) -\ln \left( \frac{\eta}{4} \right) \right] }{N}\]
\item $q$ is the $VC$ dimension $q(l_{\mathcal{H}})$ of the space $l_{\mathcal{H}}$.
\end{itemize}
\end{theo}

\noi \textbf{Proof.} \cite{Vapnik:theory} showed that, for any $\varepsilon>0$, the following inequality holds true:
\bequa
\mathbf{Pr}\left(\sup_{h \in \mathcal{H}} \delta_1[R(h),R_{emp}^{\mathbf{\Upsilon}_{N}}(h)]> \varepsilon \right)< 4\, \exp\left[\left( \frac{q \left( \ln \left( \frac{2N}{q}\right) +1\right)}{N} - \frac{\varepsilon^2}{4M}\right)N\right] \label{equa5.25}
\eequa
when conditions \textbf{C.1}, \textbf{C.2} and \textbf{C.3} are satisfied \cite{Vapnik:theory}, see inequalities 5.24 and 5.12 at pages 197 and 192 respectively). Set the right hand side of the above inequality equal to $\eta$. Then the expression of $\varepsilon$ is:
\[ \varepsilon = \sqrt{M \zeta}\]
and, therefore, from Vapnik's inequality, it follows that the inequality:
\[\sup_{h \in \mathcal{H}} \delta_1[R(h),R_{emp}^{\mathbf{\Upsilon}_{N}}(h)]< \sqrt{M \zeta} \]
holds true with probability of at least $1-\eta$.$\Box$
\\

\begin{theo}[$\mathcal{IPERM}$ applicability and VC (2)]
\label{theo5.7}
Let $\mathcal{E}=(\mathcal{T},\mathcal{OM},z,P_{z})$ be a probabilistic environment and, associated with it, a learning machine $\mathcal{LM}=(\mathcal{H},\mathcal{A})$. Let $\mathbf{\Upsilon}_{N}$ be a finite sequence of $N$ training examples from the environment $\mathcal{E}$ and $\eta$ a real number in the interval $]0,1[$. If the conditions \textbf{C'.1}, \textbf{C.2} and \textbf{C.3} are satisfied, then the $\mathcal{IPERM}$ is $\delta_2$-applicable to  $(\mathcal{E},\mathcal{LM})$ with the bound:
\bequa
\mathcal{C}=\gamma(s)\, \tau \sqrt{\zeta}  \label{equa5.26}
\eequa
where:
\begin{itemize}
\item $\gamma(s)=\sqrt[s]{\frac{1}{2} \left( \frac{s-1}{s-2} \right) ^{s-1}}$
\item The number $\zeta$ is:
\[\zeta=4 \frac{\left[ q \left( \ln \left( \frac{2N}{q}\right) +1\right) -\ln \left( \frac{\eta}{4} \right) \right] }{N}\]
\item $q$ is the $VC$ dimension $q(l_{\mathcal{H}})$ of the space $l_{\mathcal{H}}$.
\end{itemize}
\end{theo}

\noi \textbf{Proof.} \cite{Vapnik:theory} showed that, for any $\varepsilon>0$, the following inequality holds true:
\bequa
\mathbf{Pr}\left(\sup_{h \in \mathcal{H}} \delta_2[R(h),R_{emp}^{\mathbf{\Upsilon}_{N}}(h)]> \gamma(s)\, \tau\,\varepsilon \right)< 4\, \exp\left[\left( \frac{q \left( \ln \left( \frac{2N}{q}\right) +1\right)}{N} - \frac{\varepsilon^2}{4}\right)N\right] \label{equa5.27}
\eequa
when conditions \textbf{C'.1}, \textbf{C.2} and \textbf{C.3} are satisfied \cite{Vapnik:theory}, see inequalities 5.43 and 5.12 at pages 210 and 192 respectively). Set the right hand side of the above inequality equal to $\eta$. Then the expression of $\varepsilon$ is:
\[ \varepsilon = \sqrt{\zeta}\]
and, therefore, the inequality:
\[\sup_{h \in \mathcal{H}} \delta_2[R(h),R_{emp}^{\mathbf{\Upsilon}_{N}}(h)]< \gamma(s)\, \tau \sqrt{\zeta} \]
holds true with probability of at least $1-\eta$.$\Box$
\\

\noi Note that $\mathcal{WPI}$ is represented by the number $M$ in theorem \ref{theo5.6} and by the numbers $s$ and $\tau$ in theorem \ref{theo5.7}.
\\

\noi The following theorem uses a weaker $i.i.d.$ condition (\textbf{\emph{C'.3}}):
\\
\begin{theo}[Using condition \emph{\textbf{C'.3}}]
\label{theo5.8}
If the third condition \textbf{C.3} in the two previous theorems \emph{\ref{theo5.6}} and \emph{\ref{theo5.7}} is replaced by the condition \textbf{C'.3} and the two other conditions, \textbf{C.1} and \textbf{C.2} for theorem \emph{\ref{theo5.6}} and \textbf{C'.1} and \textbf{C.2} for theorem \emph{\ref{theo5.7}}, are kept unchanged, then the $\mathcal{IPERM}$ is still applicable to $(\mathcal{E},\mathcal{LM})$ with respect to the same deviation measures $\delta_1$ and $\delta_2$ and with the same bounds \emph{\ref{equa5.24}} and \emph{\ref{equa5.25}}, respectively.
\end{theo}

\noi \textbf{Proof.} To prove inequalities \ref{equa5.25} and \ref{equa5.27}, \citeN{Vapnik:Estimation,Vapnik:nature} made use of the weaker i.i.d. condition only. As a result, these inequalities remain true if condition \textbf{C.3} is replaced by condition \textbf{C'.3}. Consequently, the foregoing proofs of theorems \ref{theo5.7} and \ref{theo5.6} are still valid with condition \textbf{C'.3}.$\Box$
\\

\noi Using theorems \ref{theo5.6}, \ref{theo5.7}, \ref{theo5.8} and \ref{theo5.4}, it is now possible to develop uncertainty models for $(\mathcal{E},\mathcal{LM})$ with a guaranteed deviation $\varphi$ that is readily computable:
\\

\begin{theo}[Uncertainty Model and VC]
\label{theo5.9}
Let $\mathcal{E}=(\mathcal{T},\mathcal{OM},z,P_{z})$ be a probabilistic environment and, associated with it, a learning machine $\mathcal{LM}=(\mathcal{H},\mathcal{A})$. Let $\mathbf{\Upsilon}_{N}$ be a finite sequence of $N$ training examples from the environment $\mathcal{E}$ and $\eta$ a real number in the interval $]0,1[$. Let \hemp\ a decision rule at which the empirical risk \Remp\ reaches its minimum.
\begin{itemize}
\item If the conditions \textbf{C.1}, \textbf{C.2} and \textbf{C'.3} are satisfied, then the inequality:
\bequa
[\mathcal{D}(h_{emp}^{\mathbf{\Upsilon}_{N}},g^{\mathcal{T}})]^2  \leq  R_{emp}^{\mathbf{\Upsilon}_{N}}(h_{emp}^{\mathbf{\Upsilon}_{N}}) +  \frac{M \zeta}{2} \left(1 + \sqrt{1+\frac{4\,R_{emp}^{\mathbf{\Upsilon}_{N}}(h_{emp}^{\mathbf{\Upsilon}_{N}})}{M \zeta}} \right) \label{equa5.28}
\eequa
holds true with probability of at least $1-\eta$.
\item If the conditions \textbf{C'.1}, \textbf{C.2} and \textbf{C'.3} are satisfied, then the inequality:
\bequa
[\mathcal{D}(h_{emp}^{\mathbf{\Upsilon}_{N}},g^{\mathcal{T}})]^2 \leq \frac{R_{emp}^{\mathbf{\Upsilon}_{N}}(h_{emp}^{\mathbf{\Upsilon}_{N}})}{(1-\gamma(s)\, \tau \sqrt{\zeta})_+}   \label{equa5.29}
\eequa  
holds true with probability of at least $1-\eta$.
\begin{itemize}
\item[$\star$] $(a)_{+}=\sup(a,0)$ for any number $a\in \Re$;
\item[$\star$] $\gamma(s)=\sqrt[s]{\frac{1}{2} \left( \frac{s-1}{s-2} \right) ^{s-1}}$
\item[$\star$] The number $\zeta$ is:
\bequa
\zeta=4 \frac{\left[ q \left( \ln \left( \frac{2N}{q}\right) +1\right) -\ln \left( \frac{\eta}{4} \right) \right] }{N} \label{equa5.30}
\eequa
\item[$\star$] $q$ is the $VC$ dimension $q(l_{\mathcal{H}})$ of the space $l_{\mathcal{H}}$.
\end{itemize}
\end{itemize}
\end{theo}

\noi \textbf{Proof.} This theorem is a direct consequence of theorems \ref{theo5.8} and \ref{theo5.4}. $\Box$
\\

\noi Theorem \ref{theo5.9} establishes two uncertainty models, $\mathcal{UM}_1$ and $\mathcal{UM}_2$, for $(\mathcal{E},\mathcal{LM})$. The first one, $\mathcal{UM}_1$, is based on the weak prior information $\mathcal{WPI}(1)$ and is defined by inequality \ref{equa5.28}. The right-hand side of this inequality represents the guaranteed deviation $\varphi_{1}$ between \hemp\ and $g^\mathcal{T}$, developed on the basis of $\mathcal{WPI}(1)$. Using this function $\varphi_{1}$, the uncertainty model $\mathcal{UM}_1$ can be re-written as follows:
\bequa
\mathcal{UM}_1:\ \ \ \ \ \ \ \ [\mathcal{D}(h_{emp}^{\mathbf{\Upsilon}_{N}},g^{\mathcal{T}})]^2  \leq \varphi_{1}(N,\mathcal{H},R_{emp}^{\mathbf{\Upsilon}_{N}}(h_{emp}^{\mathbf{\Upsilon}_{N}}),\mathcal{WPI}(1),\eta) \label{equa5.31}
\eequa
with:
\begin{eqnarray}
\varphi_{1}(N,\mathcal{H},R_{emp}^{\mathbf{\Upsilon}_{N}}(h_{emp}^{\mathbf{\Upsilon}_{N}}),\mathcal{WPI}(1),\eta) & = & R_{emp}^{\mathbf{\Upsilon}_{N}}(h_{emp}^{\mathbf{\Upsilon}_{N}}) + \nonumber \\
 & & \frac{M \zeta}{2} \left(1 + \sqrt{1+\frac{4\,R_{emp}^{\mathbf{\Upsilon}_{N}}(h_{emp}^{\mathbf{\Upsilon}_{N}})}{M \zeta}} \right) \label{equa5.32}
\end{eqnarray}
The second model, $\mathcal{UM}_2$, is based on the weak prior information $\mathcal{WPI}(2)$ and is defined by inequality \ref{equa5.29}. Denoting the right-hand side of this inequality as $\varphi_2$ (guaranteed deviation developed on the basis of $\mathcal{WPI}(2)$), the uncertainty model $\mathcal{UM}_2$ can be re-written as:
\bequa
\mathcal{UM}_2:\ \ \ \ \ \ \ \ [\mathcal{D}(h_{emp}^{\mathbf{\Upsilon}_{N}},g^{\mathcal{T}})]^2  \leq \varphi_{2}(N,\mathcal{H},R_{emp}^{\mathbf{\Upsilon}_{N}}(h_{emp}^{\mathbf{\Upsilon}_{N}}),\mathcal{WPI}(2),\eta) \label{equa5.33}
\eequa
with:
\bequa
\varphi_{2}(N,\mathcal{H},R_{emp}^{\mathbf{\Upsilon}_{N}}(h_{emp}^{\mathbf{\Upsilon}_{N}}),\mathcal{WPI}(2),\eta) = \frac{R_{emp}^{\mathbf{\Upsilon}_{N}}(h_{emp}^{\mathbf{\Upsilon}_{N}})}{(1-\gamma(s)\, \tau \sqrt{\zeta})_+}   \label{equa5.34}
\eequa


\section{How to Start the Application of the Framework - Example of Wastewater Treatment Plants}

The reader is referred to \citeN{Guergachi:1} for an extensive discussion of the application of the mathematical framework developed in this paper. This section presents a very brief description of how the implementation of this framework can be started, by showing the process of defining the environment $\mathcal{E}$ of the studied engineering system. Wastewater treatment plants are chosen as an example to illustrate this implementation.
\\

\noi \textbf{Defining the Environment $\mathcal{E}_{wwt}$ for a Wastewater Treatment Plant}
\\

\noi The probabilistic environment $\mathcal{E}_{wwt}$ for a wastewater treatment plant can be an urban area, a city, a small community or a watershed. The transformer $\mathcal{T}_{wwt}$ is the wastewater treatment plant itself, which is located within the environment $\mathcal{E}_{wwt}$. This plant uses the activated sludge process to treat the wastewater generated in $\mathcal{E}_{wwt}$. The situation $\mathbf{v}$ encompasses the inputs to the plant and the state variables of the activated sludge process. It takes all its values in a space $V$. The probability density function $P_{\mathbf{v}}$ is a characteristic of the nature and amount of uncertainty associated with the environment $\mathcal{E}_{wwt}$. Two environments $\mathcal{E}_{{wwt}_{1}}$ and $\mathcal{E}_{{wwt}_{2}}$ with similar features (population, people's customs, types of industries, climate, plant configuration, \ldots) would have almost the same probability density function. The outcome $w$ is the future value of one state variable of the treatment process; it can be either the substrate (i.e., the waste) concentration  or the microorganisms concentration. The variable $w$ takes values from some subspace $W$ of $\Re$. The conditional probability density function, $P_{w|\mathbf{v}}$, of the outcome $w$ given the instance $\mathbf{v}$ is a characteristic of the plant $\mathcal{T}_{wwt}$. Two plants $\mathcal{T}_{{wwt}_{1}}$ and $\mathcal{T}_{{wwt}_{2}}$ with similar design, history, operating mode and control strategy would have almost the same conditional probability density function.

\section{Conclusions}
A mathematical framework for modeling the uncertainty in complex engineering systems is developed. This framework uses the results of computational learning theory and is based on the premise that a system model is a learning machine. A definition of an uncertainty model is given and a principle called ``Inductive of Empirical Risk Minimization'' is introduced. The applicability of this principle is examined and the concept of ``guaranteed deviation'' defined. The system model complexity is measured using the VC dimension. Based on this dimension, two different uncertainty models were developed.

\bibliographystyle{esub2acm}

%
%
%
\end{document}